# CDD-SAT IDSS

# A Real-time Cargo Damage Management System via a Sorting Array Triangulation Technique

Philip B. Alipour [1], Matteus Magnusson [2], Martin W. Olsson [3], Nooshin H. Ghasemi [4],
Lawrence Henesey [5] (*supervisor*)

[1, 2, 3, 4, 5] Blekinge Institute of Technology, Gräsvik Campus, Karlskrona, Sweden;
[1] University of Victoria, Electrical and Computer Engineering Department, Victoria BC, Canada

[1] philipbaback_orbsix@msn.com, [2] senth.wallace@gmail.com, [3] mawm06@student.bth.se,
[4] nooshin.hghs@gmail.com, [5] larry.henesey@bth.se

## Abstract

*This report covers an intelligent decision support system (IDSS), which handles an efficient and effective way to rapidly inspect containerized cargos for defection. Defection is either cargo exposure to radiation, physical damages such as holes, punctured surfaces, iron surface oxidation, etc. The system uses a sorting array triangulation technique (SAT) and surface damage detection (SDD) to conduct the inspection. This new technique saves time and money on finding damaged goods during transportation such that, instead of running n inspections on n containers, only 3 inspections per triangulation or a ratio of 3:n is required, assuming n > 3 containers. The damaged stack in the array is virtually detected contiguous to an actually-damaged cargo by calculating nearby distances of such cargos, delivering reliable estimates for the whole local stack population. The estimated values on damaged, somewhat damaged and undamaged cargo stacks, are listed and profiled after being sorted by the program, thereby submitted to the manager for a final decision. The report describes the problem domain and the implementation of the simulator prototype, showing how the system operates via software, hardware with/without human agents, conducting real-time inspections and management per se.*

## 1. INTRODUCTION

Real-time resource sorting and management with the right results on cargos, is always a challenge in critical decision makings defined within the domain of transportation industry. The time of *cargo defection/damage detection* (CDD), displaying cargo condition or its state during transport, effectively and efficiently, are hereby simulated real-time using non-human and human IDSS agents.

This project inducts a feasible technique, designed and developed from scratch, on an array of indexed cargos being shipped (or freighted) from a particular location,[1] zone or country as point *A*, thereby inspections conducted before (waiting to be loaded), during (various hazards of journey) and after transport to destination, as point *B*.

Triangulation techniques can generate real-time results on classified or indexed data representing an array of objects located in one location for a specific condition, as detected within a time limit before objects relocation. That "condition" could indicate a damaged, hazardous or any type affecting other objects and even humans negatively, if detection is not done early enough. A sorting algorithm using this "triangulation" is closely coupled with real-time agents detecting the specific condition as the mother algorithm. This algorithm is comprised of *human agents*, *non-human agents* and *sensory units* (within the agent's domain of activities) to run detection once triangulation gives them a particular index (specific object) to inspect.

---

[1] Each container is labeled in the list "{$x_1, x_2, x_3, ..., x_n$}" where $x$ denotes an alphabetical letter as the cargo column geographically occupied <u>either in a stack form or single containers</u>, and *n* is the number of the chosen containers ready for inspection.

We developed the triangulation solution on cargos via agents active in a simulation program which further attains a reliable IDSS outcome for its manager(s). The relationships of the available options coming from the IDSS system core is indicated with an ' * ' symbol, denoting that many decisions are being made throughout the conduction process of the SAT inspections via human agents. In particular, after devices detect the problem, yet the final decision depends on the inspector and the cargo manager, asking themselves what to do with the damaged version of the cargo.

## 2. DIVISION OF LABOR

We considered the contributions of our team members in developing the CDDSAT-IDSS project to be merely those allocated and assigned tasks to each member based on his/her skills that perform best at. The following solely outlines the main load of certain tasks conducted distinctly by each team member:

Dr. Lawrence E. Henesey: **Project Supervisor**

| Surname | Forename | Project Duty* |
|---|---|---|
| **Alipour** | **Philip Baback** | A-D, F-H |
| **Magnusson** | **Matteus** | C, F, H |
| **Wexö Olsson** | **Martin** | B, C, F, H |
| **Haji Ghasemi** | **Nooshin** | C, E, F, H |

**Table 1:** * Project duties for the allocated Tasks I-X: **A-** Manager, **B-** Designer, **C-** Programmer, **D-** Documenter (report), **E-** Logger (team session events), **F-** Presenter, **G-** Viewer/reviewer (proofreading), **H-** Reviser (project revisions and updates)

**Tasks done for the IDSS project:**

I. DB creation and data management issues
II. SDD Algorithm ⎤ Their conjunction forms a CDD–
III. SAT Algorithm ⎦ SAT system
IV. Connecting Tasks I, II and III: Coding to make this relationship possible
V. GUI's, portable and central communications' interface for all IDSS components of the project
VI. Bonus work or an extra addition to implement wireless communication between agents, inspectors, hardware, managers and the software
VII. DB updates supporting the interactive I/O data from/to users
VIII. Validating our output. Outputs on cargo selection, sorting and suggestions on the sorted list of damaged and undamaged cargos to the cargo manager(s)
IX. Project presentation in PowerPoint
X. Documentation, and logging

## 3. PROJECT ANALYSIS

### 3.1 BACKGROUND

The CDD-SAT IDSS project is a new system proposed to sort cargos from a population of damaged and undamaged stacks in the shortest time possible with reliable estimates [12]. These estimates are supposed to be representing the whole stack population during collaborative inspections conducted by agents for certain conditions exhibited by the containers from the cargo environment. The rules to set up this system, its objectives and the knowhow of conducting the analysis, were initially introduced in our project proposal. In this report, however, we further elaborate and demonstrate the intelligence aspects of our system based on the CDD-SAT simulation results, and real-world scenarios' dataset. The system further aims to produce such estimates to prevent further cargo damages happening in short slices of time. The system also attempts to ease any managerial decision being made on hazardous cargos to ethically save people from life-threatening situations.

### 3.2 PROBLEM DEFINITION

1. Cargo damage could happen anywhere
2. Weak intelligent support system during transport
3. Inspect and profile the damaged from undamaged
4. Parameters are defined as follows:
   – **cargo stack population** $n$,
   – **suggested containers stack for inspection** $S = (xy_i)$ from a list of stacked-up containers $\{xy_1,..., xy_n\}$

- **cargo status** or detected $D_{xyz}$ = red or orange or green
- **time to detect** $t_i$, **total time to generate status** $T_D$, **saved time** $T_{saved}$ by SAT vs. time if used by other techniques, $T_{other}$.

5. **Parametric adaptations:** $f_{local}$ = $xy_i$ coordinates for stack location, where is a local inspection made on a suggested stack of cargos by the system, *xyz* for physical volume exposures (leakages) and damaged surfaces such as punctures, holes, iron oxidation, etc. undermining any structural integrity of the containers. We adapt 3D parameters during information gather from the surfaces to 2D parameter where the stack is located from top-view during each SAT phase. This makes the sorting of the cargo array in a 2D frame much easier whilst retaining the actual 3D values intrinsically. This data is visible in terms of red, orange and green as *probabilistic P* values in the Database (DB) file. Hence, the DB component plays an important role for this dimensional values' adaptation, since we are using *a uniform language of probabilistic values* ranging from 0 to 1. Such *P* values are highlighted in the SAT system criteria in § 3.5.

The array's geometry and mathematics comply with the following computational objectives:

1. *The damaged stack in the array is virtually detected contiguous to an actual damaged cargo by software, calculating nearby distances of such cargos, delivering reliable estimates for the whole local stack population.*
2. *Relatively, a three-dimensional surface and contour mapping from containers are constructed once data is acquired at the scene. The averaged values are stored in a DB file for further processing. Parametric adaptations between xy and xyz data take place after DB access, to submit a concrete P value during value updates, giving an estimated verdict on cargo status for the whole population.*

To satisfy these objectives, we have intricately formulated the following equations in our program code to implement the SAT algorithm's logic components.

$$C_{containers} = \sqrt{n} + 1 \quad (1)$$
$$\downarrow$$
$$S_{containers} = \text{Rnd}(x, y) \quad (2)$$
$$\downarrow$$
$$D_{containers} = \text{Rnd}(r, o, g) \quad (3)$$

where $C$ is for cargo stack population *n* under triangulation defined by a square root operator with a growing limit added by 1. After finishing operations based on Eq. (1), *S* is then initiated for the suggested containers, as given by Problem Definition # 4. From there, *D* is run to detect damages from the real environment on cargos generating *rog* or red, orange, and green results. Eq. (1) is reiterated and combined with Eq. (2) to *estimate* array distances between the nearby *rog*'s as a newly estimated list of stacks located in *x* and *y* dimensions (a *j* index), or

$$(r, o, g)_{estimated} = \frac{(\sqrt{n} + 1)}{xy - x_i y_i} = j_{xy} \quad (4)$$

The deduced ratio in Eq. (4), delivers at which *xy* point, there would be a probable *r* or *o* or *g* adjacent to the actual detected ones by sensor (at least 3 containers are reported from Eq. 3). Thus, the reiteration for updating cargo results with more detections, makes the production of the *xy*-based ratios more accurate on the next trials (or phases) of SAT closing in the triangle by a factor of *xy*/2 on the grid (see Fig. 3, where the triangle shrinks by phase). This is a *SAT system evolution* as the probability over time is predicted per repeated cargo damage. This as a solution is later outlined in §3.4. Hence, with proper procedural and method callings within the code, we invoked the components of these equations to process I/O data relative to the function's side of the equation (left-hand side of the equation)

## 3.3 PROBLEM OBJECTIVES
- Detect Defected or Damaged Cargos (ranging from *low* to *severe*)
- Sort Damaged from Undamaged Cargos
- Suggest Specific Containers from the Sorted List
- Manage Cargos by human and non-human agents:
  - Detectors mounted onto the crane facility in port;
  - Detectors stationed on the vessel after cargo load
- Finalize results as a report for a Final Managerial Decision, saying that:
  - *Don't get wined up with the details, rather matches of the results in the database, inputs, outputs and updates coming from the algorithm.*
  - *Correct input gives correct output estimates probabilistically, either in form of more estimates otherwise precisions.*

## 3.4 PROBLEM SOLUTION

Our simulation method integrates **Regular and Quasi-Monte Carlo methods** [4] for computing $x_i \ldots x_n$ cargo status estimates in each triangulation phase. The median point which covers a fuzzy status i.e. "to some extent damaged" on a container stack is done via quasi-Monte Carlo method as well as using a random function as out input to inspect a stack, its surface detection points, etc. for evaluating the limits of the algorithm relative to output. The specifics of the method contributing to our solution in a nutshell are as follows:

1. We define a domain of possible inputs on $i$ as $(x_1y_1, \ldots, x_ny_n)$
2. Generate inputs randomly from a probability distribution over the domain (2D Array) giving $x_i \times y_i$,
   a. Random Function $f(n) = \text{Rnd}(i)$,
   b. We inspect cargo $i$ as a specific stack suggested by the system for inspection, for one triangle side, α, sorting the bottom row of the array (Fig. 3), for this *detection angle*,
   c. Two more $i$'s for the remaining triangle sides, β and γ: The geometry of the triangle for right and left columns sorting sides of the array is illustrated in Fig. 3.
3. We generate an estimate between the detected 3, using distances to the remaining adjacent stacks in a database
4. We aggregate and update the results from one SAT phase to another on the whole population $n$.

## 3.5 SYSTEM CRITERIA

The system displays a list of sorted cargos, color-coded and labeled for the cargo inspectors and mangers of the system. The colors on the sorted list of labels are reported and stored as **orange** (a grey state, e.g. a probability of $P = 0.2$ to $0.5$) indicating somewhat damaged cargos, **green** indicating undamaged cargos ($P < 0.2$) and **red** for significantly damaged cargos ($P > 0.5$). The inspection is conducted in three triangulation phases. Each phase refers to the sorted list either current or previously-recorded in the database to update the list as each phase successively progresses to its next for the SAT service. The probability $P$ score ranging from 0 to 1 is calculated from the results gained from the various tests performed. Here is a list of all of the important criteria defining our $P$, relative to constraints (Fig. 1):

- **Maintenance -** The detectors recheck *bounds*, and hardware status is regularly checked which is part of the monitoring process on devices such as listing any symptoms on component failures, errors in detection, etc. prior to cargo inspection.
- **Intelligence -** The intelligence is tested in a test covering several various areas:
  - *Logic*: mainly, "how the system is managed and processed?"
  - *Patterns*: imagery analysis based on plotting container surface points and measuring their distances from each other, where the program produces filtered images to identify damaged surfaces from undamaged ones (see *image binarization* from the following paragraphs). This complies with Hyosung Lee *et al.* [9] methods of computing actual damages based on surfaces and depth analysis.
- *Shapes*: Design and Graphics relative to mathematical and geometric calculation in suggesting containers as well as identifying their status.
- *Mathematics*: formulaic solutions conveying to the triangulation, and calculating graph distances between damaged and contiguously-detected damaged vs. undamaged cargos. The aim is to generate probabilistic $p_i$ results, based on "patterns" and "shapes" from above (*quite analytic* and *mathematical*), thus their sum averaged with other $p$'s, build up the final $P$ for our Q&Q analysis.
- *Physics*: detectors and sensory agents for inspection purposes employed within the cargo environment
- *Data management*: a DB raw file contains records on the sorted greens, reds and orange items
- *Knowledgebase subsystem*, based on previous SAT experience on specific data records: records on the sorted greens, reds and orange list, if the same list appear for a similar container population configuration, the results shall remain similar despite of environmental complications

- **Wireless vs. cable-based technology -** Rather than totally relying on human input, the inspection system must relay information from the cargo environment via detectors to the system, thereby processed by software for analysis and results. This is aided by using a simple wireless local area network (LAN) connection, and in worst case scenarios where atmospheric changes become dramatic, other remedies such as direct cables as well as human input (console based) are contemplated.

- **Mechanical -** The Crane Docking facility and detectors' dynamic properties involve mechanics which could affect the rate of inspections depending on what state the cargo is i.e. being unloaded, loaded or just stacked-up in some location. An optional requirement is testing mechanical skills for evaluating the time efficiency of the system.
- **Cognition -** Self and environmental awareness of the human inspector is crucial in reading the LCD unit on the detectors when they display results during detection. The validity and correctness of the results prior to the automated system relaying information to system core for processing must be monitored based on this skill.
- **Physical color sight and detection -** The color sight is tested by trying to identify cargos on colored backgrounds for a damaged section, pictured as follows: the red indicates a damaged area (in case of radiation, the physical area or material where radiation is exposed)

Data Acquisition and SAT processing conform to the following two paragraphs:

- **Fast surface-point inspection -** Simple cameras and laser-based detectors: Material surface analysis is well-explicated in the paper written by Hyosung Lee, *et al.* [9] distinguishing the damaged surface from a healthy one during camera recordings, observation and imagery analysis. The measurement satisfies the physical bounds on:

  1- Surface Roughness ,   2- Depth

  In addition, we have also prioritized an *image binarization* solution to focus on *suspicious spots* detected on the surface (a typical image capturing technique). This is to shortcut the surface damage estimates, as shown in Fig. 8, thereby conducting the 3D beam scattering technique akin to Hyosung Lee *et al.* technique [9].

- **In-depth surface point inspection -** SAT detection via sensors (CCD detectors) are thorough for an inspected unit (container) hence not a random estimate of damages. This is an accurate and efficient way to inspect a set of containers co-local to others on the vessel, or after shipping, on the platform. After image "binarization" (above paragraph) a *geological analysis tool* called Surfer 9 [10] aided us in simulating 3D and contour mapping of the captured image of the surface i.e. incoming data from the scene acquired by the detectors where cargo is located.

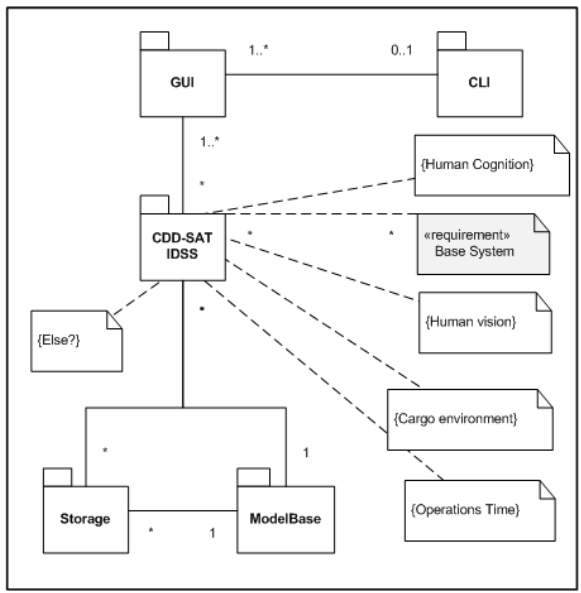

**Fig. 1:** The high-level architecture of the CDD-SAT IDSS System, classes, their relationships and constraints (the {Else?} constraint could be a potential or hidden constraint).

- **Administrator and maintenance -** System administrator must make sure all reported data, profiled information based on the DB component as well as accuracy of the devices being used by inspectors, are in position. The accuracy and correctness of data converted to interpretable data as information to the users (inspectors and managers) must be maintained by maintainers of the system on a physical level: this means that before, during, after inspections, all sensory subsystems and detectors for inspection purposes and data accuracy are checked and rechecked on a regular basis.
- **Cargo status -** Based on a sensory inspection (a detector) within the *triangle's geometric range*, [2] suggests specific containers to the inspector to inspect. The detector generates results on that container displayed to the inspector as either damaged, somewhat damaged or undamaged on his/her handheld portable detector. Note that, for an automated system, an installed rail permitting the detector to slide in-and-out, accelerates this inspection process. If a crane system is operated by an inspector, after e.g. the ship docks, the stacks of cargos are scanned via relevant detectors for defection detection.
- **Cargo records and classification -** The inspector relays data via the detector to the software for SAT estimates, once the three inspections are carried out covering the 3 sides of the triangle. The inspector's cargo is recorded after weighing the physical conditions of the cargo, estimates a list of cargos matching the description of the detected, and thus listing three containers on the record (database)[1] to produce more results. Now, the agents proceed to the next phase of the triangulation for more accurate estimates or list.
- **SAT service needs -** Informs the decision process of the inspection needs for a certain real-time occupation at cargo locations, e.g. "3 more detectors needed to inspect these cargos"
- **Agent's Interface -** A command-line interface (CLI) is a mechanism for interacting with a computer operating system or software by typing commands to perform specific tasks. Pending or in-queue detection tasks relayed from one detector (*hardware interface*) to the software DB belongs to this mechanism.

---

[2]The "range searching problems" and data structures are a fundamental topic of *computational geometry*. The range searching problem finds applications not only in areas that deal with processing geometrical data (like geographical information systems (GIS), and CAD), but also in **databases**. See also [11].

- **Human vision relative to cognition -** Based on human visual and cognitive limitations relative to the inspection environment between two containers (bearing a certain measurable distance), one must consider this inspector to fully function according to the protocols and assignments given for a cargo inspection. If issues like *dyslexia*, *color blindness* or *any human deficiencies* are involved (instantaneously, otherwise, a long-term side-effect ignored due to being a known employed worker to the company) could encounter *wrong deductions on the reading of data* as well as the quality and integrity of information process.

- **A trainable/learnable SAT algorithm -** The data is updated on regular 3-time inspections made over a population of cargos. Of course, the inspections are $3/n$ for each triangulation relative to a red-orange-green estimate result.

## 3.6 SWOT ANALYSIS

We have conducted our SWOT analysis (*strengths*, *weaknesses*, *opportunities* and *threats*) of the project and presented in the following table:

| Strengths | Weaknesses |
|---|---|
| <ul><li>Not expensive</li><li>Significant time reduction in performing cargo inspections</li><li>Significant time reduction in submitting and analyzing data in the system</li><li>Usability of the software is comprehensive and simple to run by inspectors as well as managers</li><li>Good means of developing closer integration with other partners</li><li>Scalability of cargo population for efficient inspection is unlimited</li><li>Significant cost reduction in administration and documentation (resource management) on all cargos.</li><li>The read-data off the sensors are specific and short to the system</li></ul> | <ul><li>If on-time detections for certain conditions e.g. radiation are not done, more damage on the inventory or cargos are expected (expansion of radiation exposure to the rest)</li><li>Inspectors must know and have enough knowledge to operate sensors or cargo damage readers on site</li><li>Inspectors must also coordinate with each other during each triangulation, making rest assured all inspectors have submitted their readings to the system for the next phase<ul><li>This involves human cognition and vision collaboratively (see Fig. 1 constraints)</li></ul></li><li>Can it work with other systems on the network, especially when an on-going threat is being developed?</li><li>Corner surfaces for damages from stacked-up containers might not be measurable (less evident in case of radiation)</li><li>The software generates estimated reliable profiles rather than absolutely precise profiles (we don't live in a perfect world).</li></ul> |
| **Opportunities** | **Threats** |
| <ul><li>Sensors could be run automatically with single operators (for different blocks of containers) rather than 3 inspectors for three inspections per triangle.<ul><li>This delivers real-time information</li></ul></li><li>SAT Improves great population of cargo inspections</li><li>Improve insurance and compensation on damaged cargos against savable items</li><li>Lower transaction cost for all in port community</li><li>Lower human-based activities and greater coverage of information with profiling features</li><li>Faster cargo delivery to customers due to efficient inspections and cargo profile reports</li><li>The algorithm is applicable throughout the cargo journey from source to destination: load, unload, ship, sort and delivery of containers<ul><li>The system runs the triangle's laser points (the dynamic α, β, γ coordinates at the poles) or detection system for each successive phase</li></ul></li></ul> | <ul><li>Environmentally sensitive data (activities at the site)</li><li>Requires basic IT training to operate a typical terminal for the software relative to sensors<ul><li>This is not the case if inspections are conducted through self-automation</li></ul></li><li>In case of "self-automation," devices and sensors must be checked and rechecked on the circuit's performance which imply to raw data as they read-write to the system<ul><li>Is the device still reliable i.e. is the data displayed to the inspector interpretable?</li><li>Does it require maintenance?</li><li>Does it require replacement?</li></ul></li><li>Reciprocal integrity of information sides (Input × Output = 1) must be maintained: no asymmetric or anomalies of information exhibited by the human peers, detectors and SAT software for accuracy, relative to hardware-software reliability, should be experienced. (Both sides of the equation must remain intact in terms of I/O values)</li></ul> |

**Table 2:** A representation of our SWOT analysis on the CDD-SAT algorithm.

According to our SWOT analysis, the *weaknesses* mostly lie in human errors i.e. inspectors not acting accurately and not aware of the environmental physical variables (via detectors display unit). The *strengths* emphasize the CDD-SAT software ability to analyze and estimate data effectively, efficiently and economically. From our *opportunities*, we derive that faster hardware, SAT software and less manpower plus reduced insurance viabilities, make more efficient delivery to customers. From the *threats*, we conclude that environment affects the real-time I/O data given and expected from the hardware to the user. This requires extra important attention especially if the system is self-automated (regular inspections on the operating hardware is essential).

## 4. Project Design

### 4.1 Centralized Architecture

The architecture of CDD-SAT IDSS is a layered approach with the graphical interface and console based interface on top, going down to the central part of the system that connects the storage to the model and up to the user interface. This is illustrated in Fig. 1, including constraints and a minimum requirement to run and maintain the system. This requirement is noted as a "Base System" which builds the system core and model by providing the system with relevant data to function, and in time, building information using the right interpreters from packages and programming skills. These eventually, build the knowledge-based system of CDD-SAT once implemented. To have this system built on the basis of this requirement, the necessary tools must also be incorporated to enable the system to perform CDD-SAT operations. These operations are classed as **SDD** for Surface Damage Detection, and **SAT** for Sorting Array Technique. Their association in operations, as shown in Fig. 2, builds a CDD dataset as well as results stored onto the system. The results further contribute to CDD system analysis using standard and custom functions, reference tools, libraries and packages accessible by code on the system and thus their outputs displayed on screen.

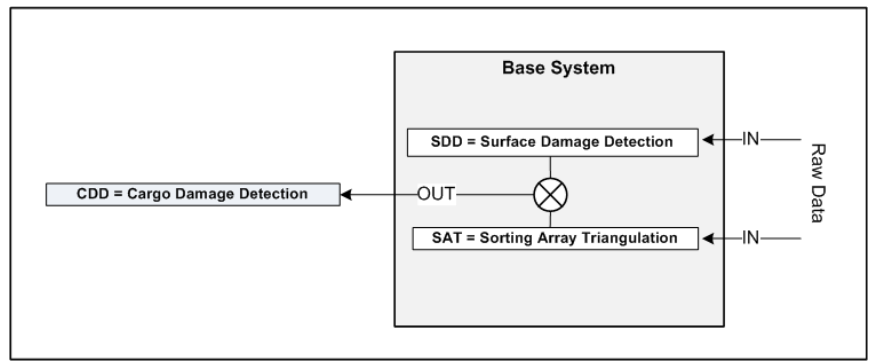

**Fig. 2**: The CDD-SAT base system. The acronym SDD stands for Surface Damage Detection, and SAT, stands for Sorting Array Technique. The junction of both results in a CDD as Cargo Damage Detection.

### 4.2 Overall Architecture

The overall architecture is illustrated in Fig. 3, where specific Use-Case diagrams with active IDSS components are also included in the design for the system. The design consists of agents interacting with the system and cargo environment where triangulation takes place. The operation is well-explicated by the following activity steps once the SAT system initiates. A layered architecture composed of use-cases is given in Fig. 4 with relevant use-case descriptions in the following subsections.

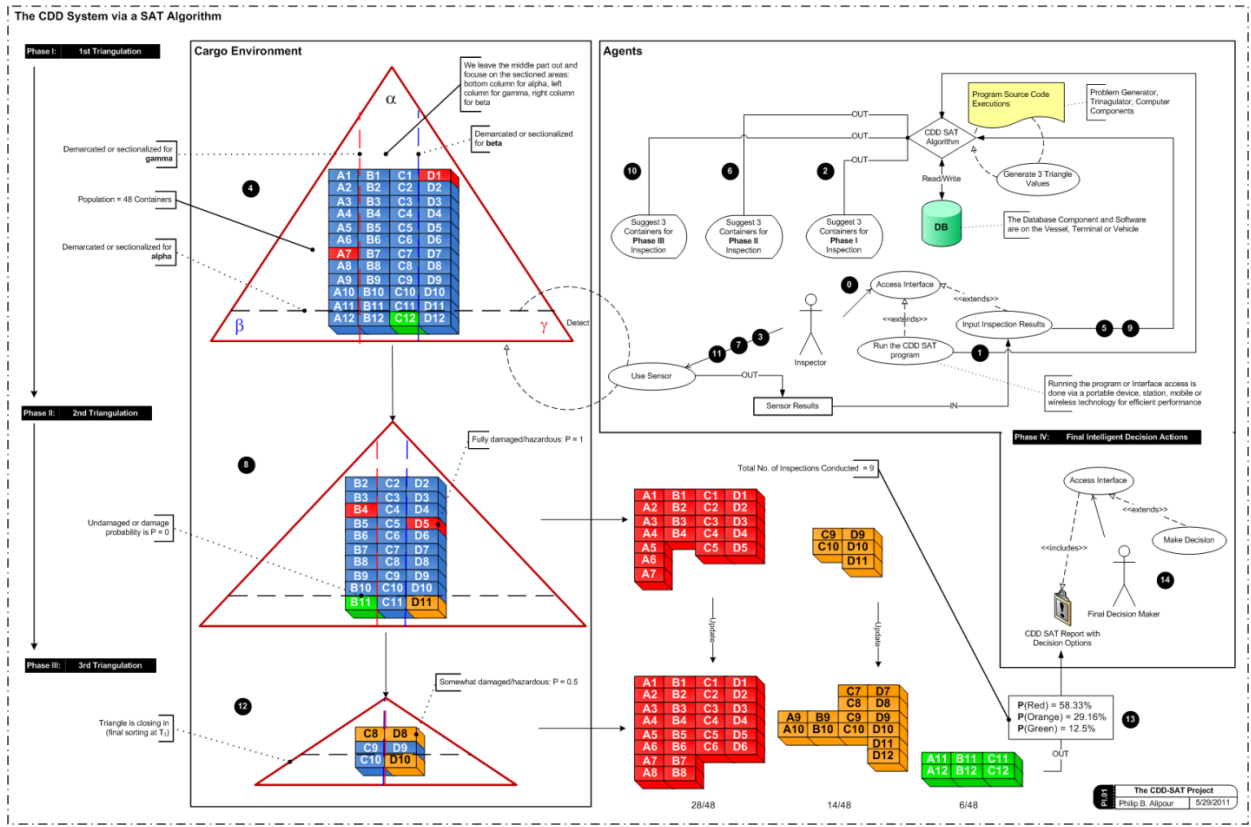

**Fig. 3:** An example of the CDD-SAT algorithm, agents, environment, the triangulation technique, software/hardware components attaining a final intelligent decision point, commencing with step 0 and ending with step 14 in four "phases" of time as well as action.

## 4.3 USE-CASE DIAGRAMS

An overview of the use-cases can be found in Fig. 4. There is also another stakeholder, namely the administrator, which is not visible in the overall plan (Fig. 4). The reason for this is that the administrator is not in direct connection with the system and the focus of this actor is on managing the database component updates, system failures and data validation on the incoming information flow concerning the status of cargos. This is not a constant check and subject to system erratic or unforeseen behavior in case of occurrence (e.g., hardware/software failures).

### 4.3.1 USE-CASE: TYPE IN CARGO POPULATION (CLI)

**Goal:**
Have the inspector input a number of stack population into the system via command-line interface (CLI)

**Preconditions:**
- Have the detectors in a functional state

**Success Condition:**
Input is being processed

**Failure Condition:**
System process failure (hardware/software) circumstantially

**Steps:**
1. Enter the *n* number representing stack population
2. Press "Enter" [↵] key

3. Population being processed by the system hardware and software SAT program
4. SAT hardware and software initiated

**Extensions:**
2.1 Cargo Triangulation is available in the system and engaged

### 4.3.2 DISPLAY SAT OPERATIONS (GUI)
**Goal:**
Have the inspector to view a suggested container to inspect by the system via graphical user interface (GUI)

**Preconditions:**
- Have the detectors in a functional state,
- Have the display unit in a functional state,
- The software is already engaged and processing and displaying suggestions

**Success Condition:**
A specific stack is displayed on screen to the user for inspection

**Failure Condition:**
System process failure (hardware/software) circumstantially

**Steps:**
1. Wait for SAT operations to finish
2. View suggested container(s) in terms of labeled stack and row # to inspect
3. Be ready to inspect the specific container using detectors

**Extensions:**
2.2 Cargo Triangulation is available in the system and engaged
3.1 The SAT program is displayed and suggests 3 specific containers to the user for inspection

### 4.3.3 RUN DETECTORS
**Goal:**
Have the inspector to inspect the suggested container and thereby processed by the system

**Preconditions:**
- Have the detectors in a functional state,
- Have the display unit in a functional state,
- The software is already engaged and process data by receiving sensory results from detectors, then display status results

**Success Condition:**
A specific stack is inspected and results are relayed to the software processor to process a SAT *red*, *orange* and *green* status outcome

**Failure Condition:**
System process failure (hardware/software), circumstantially

**Steps:**
1. Wait for SAT operations to finish
2. Press "Run Scan" button on the GUI to run detections: 2D image capture → 3D surface mapping (as first scenario) parallel to radiation unit counts if necessary (as second scenario)
3. Press "save" button or this is automatically done when "AutoSave" is checked by default
4. Wait for SAT operations to finish
5. Check results in the profile as red, orange or green
6. Be ready to inspect another specifically suggested 3 containers using detectors for the next phase

**Extensions:**
Cargo Triangulation is available in the system and engaged
2.1 The SAT program is displayed and processes 3 specific containers from the user as inspected by the surface map processor of the program:
   a. XY parametric data (*location*) is used by the software to generate estimated cargo status results on other localized containers *by the program*
   b. XYZ parametric data (*surface mapping 3D-points*) is used to compute the actual condition of surfaces captured *by the camera*

c. X parametric data (*radiation units*) is used to compute the degree of exposure to radiation in case of radiation detected by a *Geiger-Muller Counter*

3.1 Store/update data by the program

3.2 Add/update inspection by the program

6.1 Three <u>detections</u> (α, β, γ detectors) could run <u>concurrently</u> for each SAT phase, reducing the time factor down to a third if proper parallel coding is in place. Choose "*concurrent scan option*" radio button, then click on "Run Scan".

6.2 A regular manual scan mainly involves one detector at a time (this is the user's default mode). The "*step-by-step or sequential scanning option*" is checked by default, the user clicks on "Run Scan".

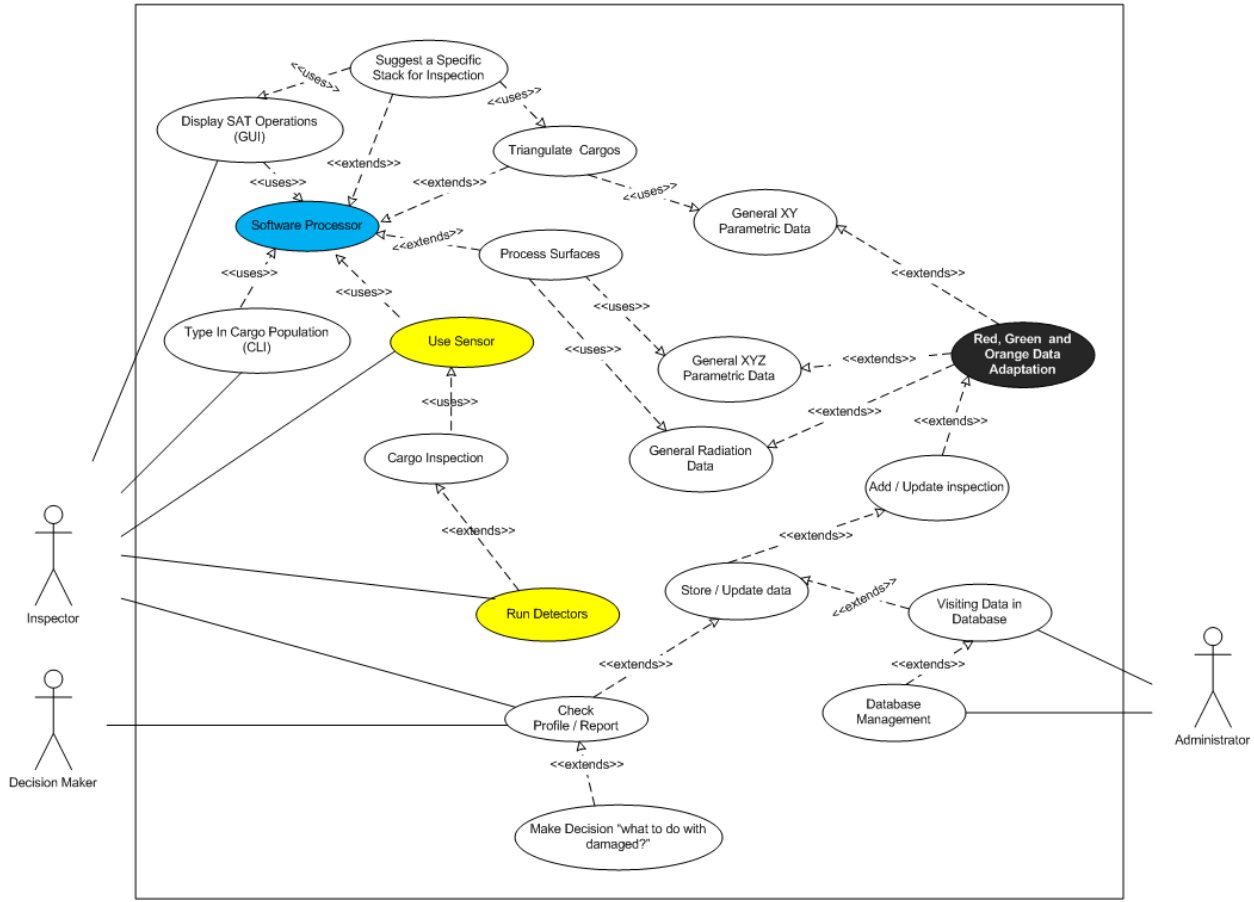

**Fig. 4:** An overview of the use-cases: yellow cases represent a hardware property used by an actor; blue case is a combination of hardware and software interfacing other cases; the black case represents "parametric adaptation" based on generated results by the system; white cases represent an extended or usable case by an actor or system components.

### 3.2.1 CHECK PROFILE
**Goal:**
Have the inspector and/or decision maker to check cargo profile generated by the system after *k* phases of inspections (the reiteration of Cases 4.3.2 and 4.3.3) for all sorted cargo population

**Note:** The phases are finished once the stacks are completely processed i.e., 3 inspections per phase, and sorted by the program after all successive phases of inspections are finalized.

**Preconditions:**
- Have the display unit in a functional state,

- The software is already engaged, processing and delivering estimates on the whole cargo population status

**Success Conditions:**
- The data is reported to the inspector and manager for making a decision between damaged and undamaged goods,
- The validity of data updates is additionally checked by the administrator if necessary

**Failure Condition:**
System process failure (hardware/software), circumstantially

**Steps:**
1. Wait for SAT operations to finish
2. Check final results in the profile as red, orange or green population
3. Make decision on the damaged and undamaged cargos as listed by the system

**Extensions:**
Cargo Triangulation is finished in the system and damaged vs. undamaged cargos are reported
2.1 Visiting data in the database by the administrator
   a. Managing the database
2.2 Manager makes decision what to do with damaged cargos based on the profile results
2.3 Inspector checks the profile and in case, runs secondary inspections

## 4.4 CLASS DIAGRAMS

The collection of the class diagrams for the different packages in the system along with a short description of them is given in Fig. 5.

Since the system included so many variables and instantiations to fulfill the requirements of the system functions, we merely listed the main players possessing certain roles in establishing a relationship between one UML element and another. Therefore, illustrating a concrete set of diagrammatic representations on classes, components, packages drawn between SDD and SAT systems, their subsystems' functionalities, processes and operations within the CDD-SAT system, was the most conceivable approach. Diagram (Fig. 5), contains all the collaborations of every class created in VB, C++, and Java, pertinent to our pseudocodes (§§ 5.1.1 and 5.1.2), satisfying system objectives (§ 3.3) to constitute the CDD-SAT hybrid simulator. The VB, C++ API interfaces with both *specialization and generalization relationships* (*reciprocal* and *hierarchical*) to other UML elements of the CDD-SAT system are highlighted in the diagram. Moreover, the Java Applet class plays the final role in displaying all the collated, processed and disseminated data from the previous CDD-SAT phases of the simulation in form of tables plus a *time savings* graph on screen (later given as Eq. 5). The processors system and subsystem deal with Input/output signals, from the moment where cargo surface data are acquired from the system, until the moment where processes reach conclusions and verdicts portraying the cargo status in the DB component. Of course, the 'repository system' acts as a middleware to other systems, preserving a specific format during I/O data transmissions. The format is based on the storage package which precludes any noise-form structure or sequence of raw 0 and 1 data. In fact, the formatted repository preserves the *piped* and *packeted* digital forms of data as other systems' meaningful output into the database. Therefore, the information enabling the user as a central link to communicate between all SDD and SAT systems during communication is done via signals, obeying the DB assigned protocol within the processors subsystem.

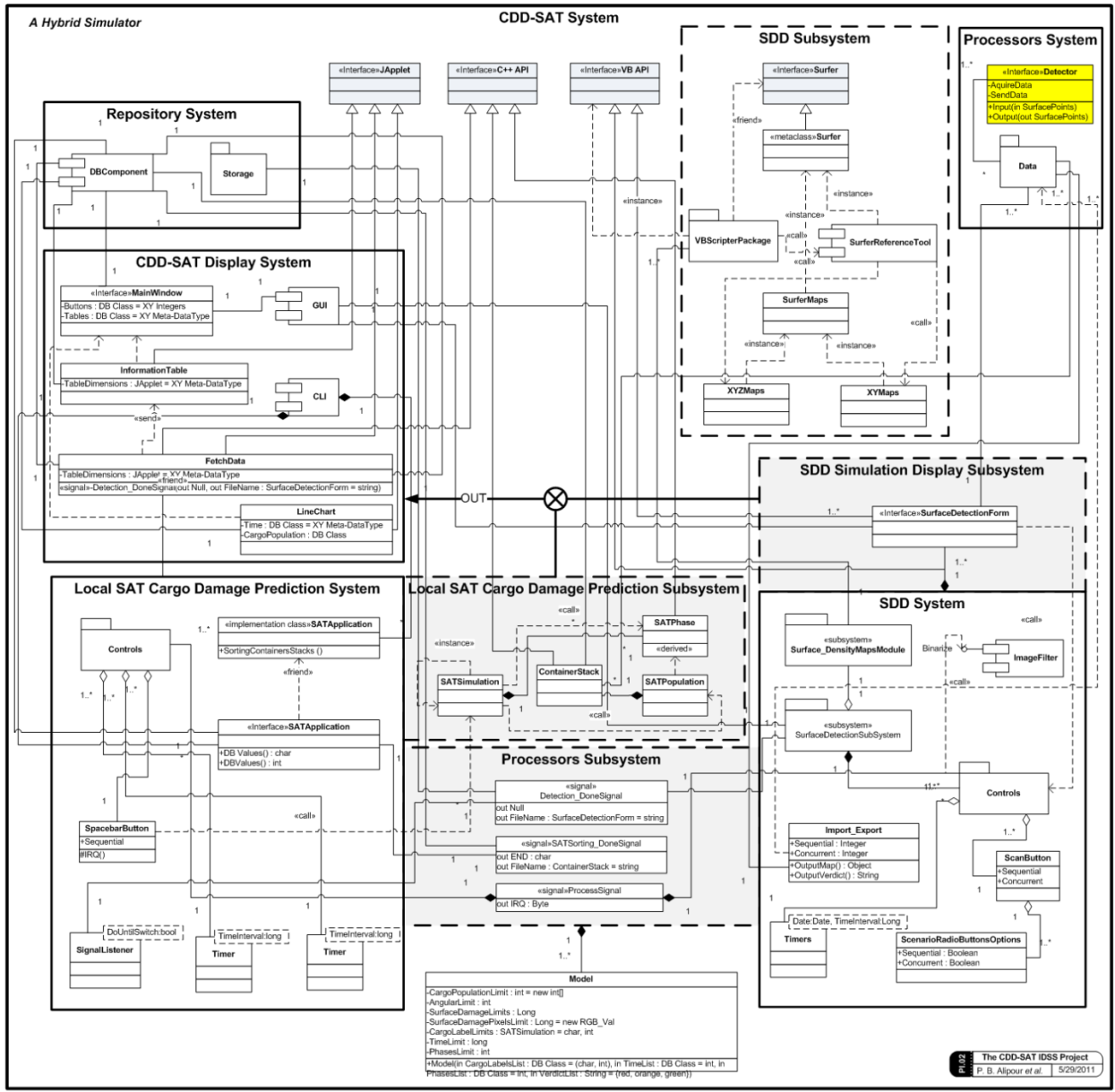

**Fig. 5:** An overview of CDD-SAT UML elements: classes, components, packages used within the systems, their relationships, functions, operations and attributes, building the hybrid CDD-SAT simulator. The hardware component is highlighted in yellow; subsystems in grey; top classes as API's and Applet colored in light blue.

The assigned protocol is a customized agreement of how data should be stored, and thereby programmers have a certain standard to make their program communicate with other programs during operations. The current version of the protocol where all subsystems communicate successfully from one CDD-SAT phase after another is shown in Fig. 7 of our DB section.

The 'CDD-SAT display system' is the resultant of the *summing junction* (association) of both SAT and SDD systems according to Fig. 5, (recognizable by the $\otimes$ symbol). Meaning that, the two SDD and SAT systems communicate with each other, after and during, the occurrence of SAT and SDD phases of the algorithm. The results and operations are displayed on screen mostly via GUI, and in the background, the CLI interface. The CLI receives input right at the beginning from the user indicating the cargo stack population ranging from 3 to $n$ stacks. So, a stacks' input of $n < 3$, is not acceptable to the systems, reasoning that there is at least 3 containers or stacks to be inspected by detectors, concurrently (per phase).

The 1 or 2 containers input, given by the user in this exception, will not be processed in terms of SAT, since SAT is a sorting array system dealing with an $n$ cargo population. However, if the system only runs on the basis of SDD, which analyzes a surface ranging from 1 to a set of containers ($n$), this would be acceptable.

In case of inspection of one container only, and not the population of stacks, the users disable the SAT option from the software, and merely do their analysis by the SDD part of the IDSS system. Therefore, in this case, no errors are reported.

The hybrid simulator, according to Fig. 5, clearly demonstrates how all interfaces via GUI and CLI components are in real-time and concurrent communication. Indeed, there is a "model" class sitting right at the bottom with a strong composition to the incoming and outgoing signals occurring between interface controls (program timers), scenario closure points, and their calculations on the stack population, time, surface damage pixels, and angular, etc.-limits of the program, defining the parameters of the CDD-SAT system.

All *detailed descriptions* for each UML element are provided in our documentation located in a "/doc" folder, fulfilling each coded solution on the system, and submitted as a supplementary document to the prototype, on its technical account.[3] Nonetheless, the brief description of our hybrid simulator elements such as classes, components, and their roles in Fig. 5, is given by Table 3.

The conversion of data types is exemplified in our classes, e.g., OutputMap from a VB class for each integer data type displayed, analyzed (compared with other points) and stored as an *xy* point of an *xy* topological map, otherwise an *xyz* reference point to an *xyz* topological map, is either exported in integer, to form a grid map, otherwise, a string to store a *verdict* {red, orange, green} in the database file based on "SurfaceDensityMapsModule" processes and analysis. The "module" calls specific map tool instances from packages, in this case, Surfer 9 (or 12) to satisfy *xy*, and *xyz* object map analysis. The decomposition of Surfer 9 [10] and other applications, their interfaces (API), class instances, components, packages communicating via signals occurring between interfaces, has been shown in Fig. 5. The Surfer program is treated as a reference tool, and by using a script language like VBScript, importing and exporting values on maps by the "module" becomes feasible.

Finally, the results are published online (wireless technology from § 3.5) to the inspector and cargo manager. Here, the Java Applet *browsing solution* for displaying DB results and time savings (performance) gets relevant to the CDD-SAT architecture.

For each VB and C++ API or Java Applet interface, we have signified the very notation of how to exist in programming terms when one demonstrates his/her skills between high-level OOP and low-level machine language? Of course, the answer is obvious.

The conversion of any data type is conditional to the way the code is compiled prior to implementation. There is a difference in how we interpret/compile the words, expressions, etc. in binary to fulfill our class relationships.

---

[3] Ask authors for the full supplementary documentation of the codes on the prototype.

| Element Name | UML Element : Domain Interactions | Description | Programming Language |
|---|---|---|---|
| **Detector** | Class <Interface>: **Processors System** | A sensory device used for real-time data acquisitions from the cargo environment, and relays the same data to the software | N/A |
| **Data** | Package <system>: **Processors System** | Raw I/O data is located in this package | N/A |
| **SurfaceDetectionForm** | Class <Interface>: **SDD Display System** | The main interface with buttons, picture boxes, for SDD verdict and analysis | VB |
| **Surface_DensityMaps** | Module <subsystem> : **SDD Subsystem** → *Surfer* → **SDD System** | Invokes methods in another program called Surfer 9 as a topographical reference tool. It processes data on imported *xy* and *xyz* maps in form of density (contour) and 3D surface maps | VBScript |
| **SurfaceDetectionSubSystem** | Subsystem<subsystem>: **SDD System** | Comprises of controls and other objects of the SDD program, operating in both, foreground and background. | VB |
| **Controls** | Package : *SurfaceDetectionSubsystem* : **SDD System** | Consists of other classes to scan and process imagery data inclusive of timers to time the length of the processes | VB |
| **Timers** | Parametric Class : *Controls Package*: **SDD System** | Part of the [controls] package to run and measure time on sequential and concurrent scans i.e. **TotalDetectionTime** = T seconds stored in the DB file | VB, VBScript |
| **ScanButton** | Object Class: *Controls Package*: **SDD System** | Part of the [controls] package to run sequential and concurrent scans by user's choice | VB |
| **Import_Export** | Class: *Data Package*: **SDD Processors System** → **SDD System** | Imports and Exports relevant data as an imagery type depending on either concurrent or sequential mode; it outputs the verdict as {red, orange, green} | VB, VBScript |
| **Surfer** | Class <metaclass>: **SDD Subsystem** | A 2D-3D simulation toolkit with VBScript package support for processing *xy* and *xyz* maps | VBScript |
| **SATApplication** | Class <implementation class>: **SAT System** | The main application to implement and run SAT by processing values from the CLI component (arguments) and DB values | C++ |
| **SATApplication** | Class <Interface>: **SAT System** | The main application to interface values by generating them to the DB for other parts of the system to access | C++ |
| **Controls** | Package : **SAT System** | Consists of other classes to generate and sort cargo stack array, inclusive of timers to time the length of the processes | C++ |
| **Timer** | Parametric Class : *Controls Package*: **SAT System** | Part of the [controls] package to run and measure time of the processes occurring within the SAT System i.e. **TotalSortingTime** = T seconds from the **SATApplication** stored in the DB file | C++ |
| **Timer** | Parametric Class : *Controls Package*: **SAT System** ← **SAT Subsystem** | Part of the [controls] package to run and measure time of the processes occurring within the SAT Subsystem **TotalSortingTime** = T seconds from the **SATSimulation** stored in the DB file | C++ |
| **SignalListener** | Parametric Class : **SAT System** ← **SDD System** | It waits for a signal to be received by the **SATApplication** program, coming from the **SDD System** | C++ |
| **SpacebarButton** | Class : *Controls Package*: **SAT System** | A class that passes an IRQ signal to the system i.e. via a standard spacebar function to invoke a phase or step of the simulation | C++ |
| **SATSimulation** | Class: **SAT Subsystem** ← **SAT System** | Instantiates Simulation Sorting Components (graphical containers + | C++ |
| | | labels) based on calling **SATPopulation** derived from a **SATPhase** (1 to *k* phases) | |
| **SATPhase** | Class: **SAT Subsystem**← *DB Component*: **Repository System** | Instantiates phases of the system engaging Controls from **SAT System** and **Processors Subsystem** via **SATSimulation** class | C++ |
| **SATPopulation** | Class <derived from SATPhase>: **SAT Subsystem**← *DB Component*: **Repository System** | It populates containers and relative graphical dimensions based on DB values read by the **SATSimulation** via **ContainerStack** class | C++ |
| **ContainerStack** | Class: **SAT Subsystem**← *DB Component*: **Repository System** | Instantiates graphical stacks of containers and reconfigures them by accessing the DB component | C++ |
| **Detection_DoneSignal** | Signal: **SDD System** → **Processors Subsystem** → *Storage Package*: **Repository System** | Sending a signal as a message stored in a specific location in the repository system for other programs that this job is done (in this case, SDD operations) | VB |
| **SATSorting_DoneSignal** | Signal: **SAT System** → **Processors Subsystem** → *DB Component*: **Repository System** | All SAT sorting, data value generations and I/O read and write is finished by the **SATApplication** and the termination of program is stored with an "END" message in the DB file (see Fig. 7) | C++ |
| **ProcessSignal** | Signal: **SAT Subsystem** → **Processors Subsystem** → **SDD System** → **VB** *Controls* → **Processors Subsystem** | Typical button strike functions passed by the **SATSimulation** program relative to SDD System controls during each simulation and detections phases | C++, VB |
| **mainWndow** | Class <interface>: **CDD-SAT Display System** ← *DB Component*: **Repository System** | The main Window with buttons and tables | Java |
| **infoTables** | Class: **CDD-SAT Display System** ← *DB Component*: **Repository System** | Responsible for drawing tables and columns with related data | Java |
| **fetchData** | Class: **CDD-SAT Display System** ← *DB Component*: **Repository System** | This class works with DB file: fetches and processes data | Java |
| **lineChart** | Class : **CDD-SAT Display System** ← *DB Component*: **Repository System** | It draws the chart. The vertical axis represents "Time" while the horizontal axis represents "cargo population" | Java |
| **Model** | Polymorphic Class | All maximum and minimum limits of the operations conducted between SAT and SDD systems are defined within the attributes and operations of this class. All classes are instantiated either in VB, C++ or Java in this hybrid simulator | VB, C++, Java |

**Table 3:** represents all classes' specifications on Fig. 5, their domain relationships within their systems and subsystems. For example, the notation **SDD Subsystem** → *Surfer* → **SDD System** for a class denotes that the subsystem implies to the domain of the system via a class called "Surfer" called by a module named "Surface_DensityMaps".

On the other hand, we are not bound to delimit ourselves to objective programming by some language in our selection. This is quite visible when we process, analyze data, as an information class, converted to a knowledge-based system, otherwise, to be ignored due to not being robust enough in collaboration with program classes in operations (languages listed in Table 3).

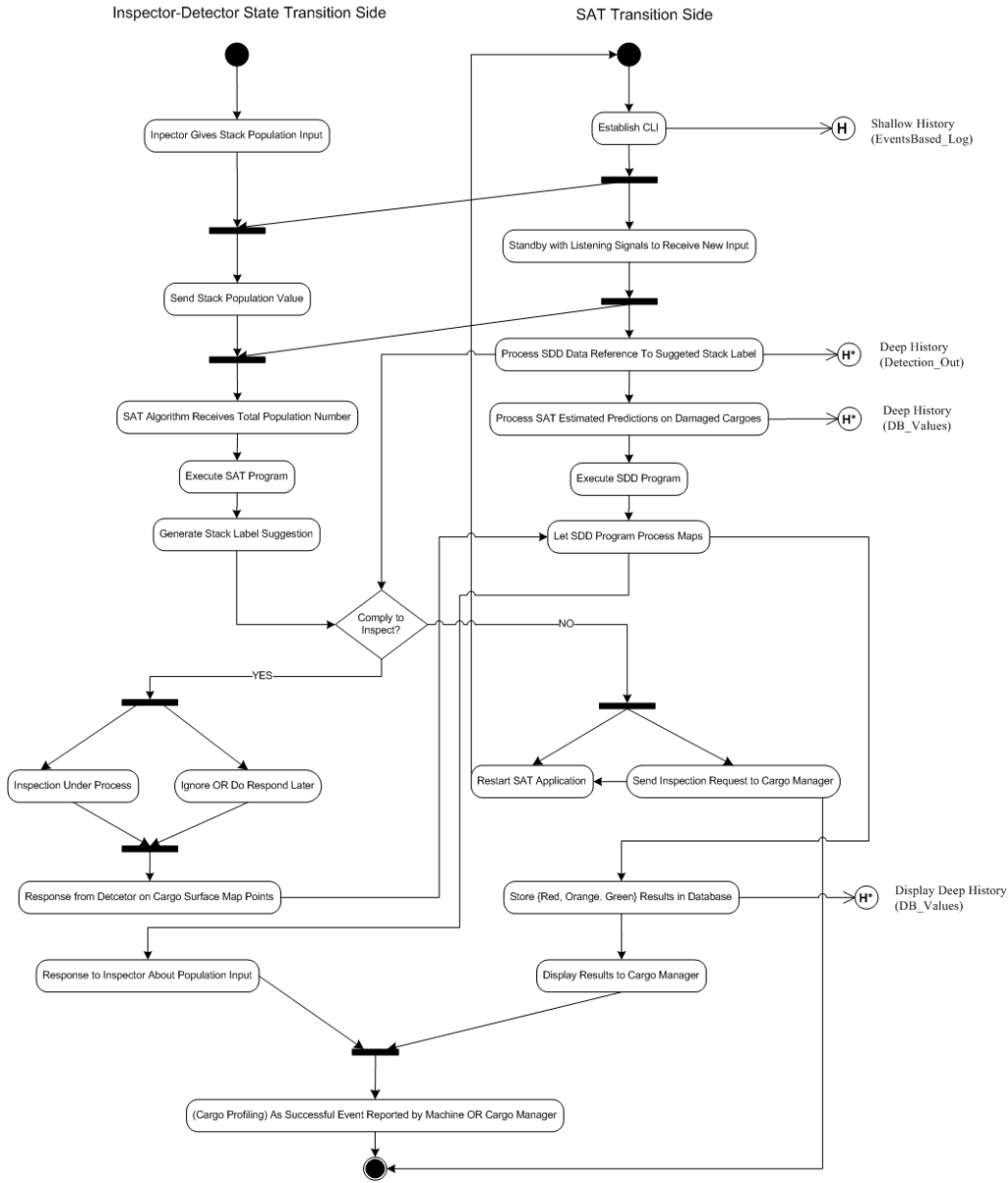

**Fig. 6:** An overview of CDD-SAT Activity Diagram. The system synchronizes between SDD and SAT states. Therefore, the conjunction of all states shall remain lucid in all UML operations i.e. inspector and the program, from the point of specific cargo inspection suggestion down to the results in the profile generated by the program.

## 4.5 ACTIVITY DIAGRAM

In Fig. 6, the flow of how the inspector conducts an SDD relative to SAT phases throughout the system is illustrated, and ends by a successful profiling of the cargos.

## 5. Project Implementation

As listed in Table 3, our prototype was implemented in a hybrid fashion i.e. being coded in different programming languages supporting *software framework interoperability* [5], where the team members were most confident and able to effectively perform their tasks in their preferred language. This allowed all members to play along the IDSS requirements fulfilling the objectives of the project as well as I/O data quality (avoiding any *garbage-in garbage-out*, GIGO, encountering). This maintains good performance on the simulation part of the project. A well-defined strategy was taken into action between our developed samples and certain portions of the prototype. For instance, the main GUI programmed in C++ for the user satisfying the graphics of the application, was partly a sub-GUI programmed in classic VB and VB.NET for the inspector running the detectors. The good thing about .NET framework is that "it includes a large library and supports several programming languages, which allows language interoperability (each language can use code written in other languages)" [5].

The simulation background however, was initially programmed in C++ or C-compatible Win32.exe console based (the CLI class, Fig. 1 or 5), which processed the main mathematics anchored behind the SAT model on cargos (see § 5.1.1 pseudocode). All data acquisitions and post-processes were, recorded in one raw DB file representing two corresponding DB files as exemplified in the following tables. All data were recorded according to an agreed custom format between team members. These data built our *central information* and at each team member's disposal during I/O data access, read and write operations. We assigned one standard protocol/format to our database files (Tables 4 and 5) in our implementation. This enabled us to refer to the database and send relevant signals to each other through creating new temporary files in aim of proceeding from one SAT phase to another. Thus, we practically created a *hybrid simulation* comprised of executable subprograms triggering one another for each contributing scenario. The chronology of events was never disrupted based on this hybrid development as far as we disciplined ourselves to stick to the agreed DB management standard, corresponding files during data transactions accessible by our codes.

**Table 4** below, represents the I/O data before sorting, during, and after sorting (cargo list and cardinality results) for an example of 48 stacked-up containers (just like the diagram). The minimal input is the cargo population, in this case, 48 containers highlighted in black.

|  | Total Container Population | Sorting Population for Alpha | Sorting Population for Beta | Sorting Population for Gamma | Detected Alpha | Detected Beta | Detected Gamma | List of Red | List of Orange | List of Green | Sorted Population | Inspections |
|---|---|---|---|---|---|---|---|---|---|---|---|---|
| Phase 1 | **48** | 4 | 12 | 12 | C12 | D1 | A7 | A7, D1 | 0 | C12 | 28 | 3 |
| Phase 2 | 30 | 3 | 10 | 10 | B11, D11 | D5, D11 | B4 | B4, D5 | D11 | B11 | 23 | 3 |
| Phase 3 | 6 | 2 | 3 | 3 | D10 | D8, D10 | C8 | 0 | C8, D8, D10 | 0 | 8 | 3 |
| Total |  |  |  |  |  |  |  |  |  |  | 59 | 9 |

**Table 5** below, mainly represents lists of sorted and 'to-be-sorted' containers corresponding to the relative attributes (columns) of Table 4, from phase 1 to phase 3, whereas phase 3 submits the final estimated results to the manager at phase 4.

|  | Sorting Population for Alpha | Sorting Population for Beta | Sorting Population for Gamma | Est. List of Reds | Est. List of Orange | Est. List of Greens | Sorted Population by Ratio |
|---|---|---|---|---|---|---|---|
| Phase 1 | A12, B12, C12, D12 | D1, D2, D3, D4, D5, D6, D7, D8, D9, D10, D11, D12 | A1, A2, A3, A4, A5, A6, A7, A8, A9, A10, A11, A12 | A7, D1 | 0 | C12 | 0.0625 |
| Phase 2 | B11, C11, D11 | D2, D3, D4, D5, D6, D7, D8, D9, D10, D11 | B2, B3, B4, B5, B6, B7, B8, B9, B10, B11 | A1, A2, A3, A4, A5, A6, A7, B1, B2, B3, B4, C1, C2, C3, C4, C5, D1, D2, D3, D4, D5 | C9, C10, D9, D10, D11 | B11 | 0.5625 |
| Phase 3 | C10, D10 | D8, D9, D10 | C8, C9, C10 | A1, A2, A3, A4, A5, A6, A7, A8, B1, B2, B3, B4, B5, B6, B7, B8, C1, C2, C3, C4, C5, C6, D1, D2, D3, D4, D5, D6 | A9, A10, B9, B10, C7, C8, C9, C10, D7, D8, D9, D10, D11, D12 | A11, 12, B11, B12, C11, C12 | 1 |
|  |  |  |  |  |  | 28, 14, 6 |  |

## 5.1 Simulator Pseudocodes

The following pseudocodes were implemented in our program, initially written by Philip and distributed to all team members for implementation. The graphics issuing lines of code in §5.1.1, employed certain design components for constructing the GUI, and are initially inherited from Fig. 3 plan. The I/O background operations representing the SAT Pseudocode lines # 12 to 87 were coded by Martin and Matteus. Database configuration, standardization and protocol results assigned to all of the team members to conduct the right standard, were further coded and displayed by Nooshin, where she also employed the relevant plug-in in Java to run Excel (spreadsheet) in generating the time of effort or the time saving point graphs. CAD Pseudocode, the hybrid CAD-SAT code successions and evaluation, detection codes proportional to DB I/Os assignments were done by Philip. The VBScript code written by Philip, focused not only on implementing the binarization filter on imagery data, also 3D surface and density mapping via Surfer 9 for the *xyz* coordinates with relative *P* outcomes (damaged *vs.* undamaged), whilst *xy* local array sorting being supplied by Martin and Matteus's codes using the same *P* to estimate more subsequent *P*'s through the database.

The scenario-based outcome of the pseudocode implementation is given in §§ 5.2 , 5.3 and 5.4.

### 5.1.1 SAT Pseudocode

**The CDD-SAT pseudocode:** *The following lines of pseudocode denote a hybrid representation of an actual code for any applicable programming language prototyping this algorithm:*

***Cargo Array Triangulation Algorithm***

```
START
1.  /* ********************************* */
2.  /* Created by: Philip B. Alipour
3.  /* Coded in C++ and Java languages by:
4.  /* Matteus Magnusson, Martin W. Olsson,
    /* Nooshin H. Ghasemi
5.  /* Created on August 31, 2011
6.  /* Last update: May 24, 2011
7.  /* ********************************* */
8.  /* Pre-settings, graphics and I/O /*
    variables of the program          */
9.  /* ************************** */
10. #Include necessary library files and/or
    packages
11. …
12. /* MAIN PROGRAM 1: */
13. Create the main graphics and I/O
    function() { //relevant to the following
    lines, create sub-functions with or
    without arguments, and call them when
    necessary
```

14. Get user input for n as Cargo Population; //this is when the user wants to know the cargo status based on some reason: environmental, event, regular inspection, etc.
15. FOR i = 1 to n {
16. Instantiate a 2D array of containers;
17. Fill 2D empty space with graphical containers; //e.g. in shape of a set of rectangles
18. Label containers as Ai for rows of column #1, Bi for rows of column #2, Ci for …, AAi for …,
    ABi for …, etc.; //we concatenate character A with i to generate Ai, where this applies to the rest in an enumerative fashion or as i grows
19.    }
20. Sort graphical containers for Object Defection Detection in three sections on the 2D array = {alpha, beta, gamma};
21. Instantiate Colors set = {Red, Green, Orange}; //use a standard RGB package
22. Instantiate Random Function f = Rnd (i);
23. Generate values of $f_i$ for {alpha, beta, gamma} on screen; //program suggests container number $f_i$ to be inspected by user
24. Get user or detector input for {alpha, beta, gamma}; //detection or sensing device delivers the output to user and database
25. Store {alpha, beta, gamma} results in a file or database;
26.   }
27. /* END OF MAIN PROGRAM 1 */

28. /* ************************************ */
29. /* Cargo Array Triangulation (CAT) sub-algorithm starts from here */
30. /* ************************************ */
31. Create a CAT function(){
32. IF alpha list = {$f_1$, $f_2$, $f_3$, $f_4$, $f_5$, …, $f_n$} AND $f_i$ is Red THEN //maximum is n, i is the randomly assigned identified object
33. Print positive defection for $f_1$ to $f_i$; //probable defection is 1 in this specific range
34. Color the $f_i$ container Red;
35. ELSE IF alpha list = {$f_1$, $f_2$, $f_3$, $f_4$, $f_5$, …, $f_n$} AND $f_i$ is Green THEN
36. Print negative defection for $f_1$ to $f_i$; //probable defection is 0 in this specific range
37. Color the $f_i$ container Green;
38. ELSE IF alpha list = {$f_1$, $f_2$, $f_3$, $f_4$, $f_5$, …, $f_n$} AND $f_i$ is Orange THEN
39. Print somewhat defection for $f_1$ to $f_i$; //probable defection is 0.5 in this specific range
40. Color the $f_i$ container Orange;
41. Store All Printed Values into a Database
42. END IF
43. IF beta list = {$f_1$, $f_2$, $f_3$, $f_4$, $f_5$, …, $f_n$} AND $f_i$ is Red THEN
44. Print positive defection for $f_1$ to $f_i$;
45. ELSE IF …
46. …
47. END IF
48. IF gamma list = {$f_1$, $f_2$, $f_3$, $f_4$, $f_5$, …, $f_n$} AND $f_i$ is Red THEN
49. Print positive defection for $f_1$ to $f_i$;
50. ELSE IF …
51. …
52. END IF
53. }
54. /* END OF CAT FUNCTION */

55. /* MAIN PROGRAM 2: */
56. Create the main CAT calculations function(){ //step 2 and step 3 triangulations
57. Read Database Values;
58. IF database empty THEN
59. Set a step variable value s = 0; //no triangulation steps executed
60. Exit and execute MAIN PROGRAM 1;
61. ELSE
62. Set s = 1; //step 1 is over
63. FOR the detected $f_i$ to $f_n$ of the array {
64. Clear $f_i$ to $f_n$ containers from the graphics array or simulation platform;
65. Display them as a separate group of containers; //red in the group of reds, green in the group of greens and Orange, in the group of orange
66. Reiterate Triangulation starting from the $f_2$ to $f_{n-1}$ of the new columns = B(i+1), C(i+1), D(i+1) to D(i-1), C(i-1), B(i-1); //this applies to other labels e.g. AAi, ABi, ACi, etc.
67. }
68. Update Array Configuration, {alpha, beta, gamma} list, Color based on triangulation results;
69. Set s = 2; //step 2 is over
70. END IF
71. IF s = 2 THEN
72. FOR the remaining $f_i$ to $f_n$ of the array {
73. Reiterate Triangulation starting from $f_i$ to $f_{n-2}$ of the new columns = Ci, Di to D(i-2), C(i-2);
74. }
75. Update Array Configuration, {alpha, beta, gamma} list, Color based on triangulation results;
76. Set s = 3; //step 3 is over
77. ELSE IF s = 1 THEN
78. Exit and GOTO line # 57;
79. ELSE IF s = 0 THEN
80. Exit and execute MAIN PROGRAM 1;
81. END IF
82. Compute the total number of positively detected objects;
83. Compute the total number of negatively detected objects;
84. Compute the total number of partly positively detected objects;
85. Print list of all sorted objects in Database;
86. Generate Message with Options for sorted cargos = {reject, isolate, control, accept, else};
87. /* END OF MAIN PROGRAM 2 */

**HALT**

### 5.1.2 CDD PSEUDOCODE
### Cargo Defection Detection (CDD) Algorithm

Detection via a specific sensing device for a specific situation starts here

```
START
1.  /* ********************************* */
2.  /* Created by: Philip B. Alipour
3.  /* Coded in Classic VB, VB-Script and
4.  /* VB.NET by: Philip B. Alipour
5.  /* Created on August 31, 2011
6.  /* Last update: May 24, 2011
7.  /* ********************************* */
8.  /* Pre-settings, graphics and I/O
    variables of the program         */
9.  /* ********************************* */
10. #Include necessary library files and/or
    packages like Surfer 9
11. …
12. /* MAIN PROGRAM 0: */
13. Create the main detector function() {
    //relevant to the following lines, create
    sub-functions with or without arguments,
    and call them when necessary
14. IF Shape Match = Ordinary Container AND
    Right Cargo Shape Volume Dimensions THEN
15. Ignore further detection;
16. Generate Profile as Undamaged Container
    → Cargo → Item(s);
17. ELSE
18. Analyze for Damages OR Hazardous OR
    Radiation Type Objects{
19. Run Detection Mode via a Geiger-Muller
    Counter on Radiation;
20. IF NOT
21. Run Detection Mode for Damaged Objects
    using 2D-3D Surface, Binarization and
    Contour Map Analysis + OCR real object to
    modifiable text analysis on damaged/hazardous
    listed and non-listed objects in the
    cargo; }
22. Generate Profile as Damaged Container →
    Cargo → Item(s);
23. Generate Final Profile After Value
    Comparisons;
24. Generate Options and Decide {
25. Generate Message with Options for sorted
    cargos = {reject/isolate/control, accept,
    else} }
26. } /* END OF MAIN PROGRAM 0 */
HALT
```

## 5.2 DATABASE

Resource sorting and management with the right *P* results on cargo status are situated in the DB component where the hybrid simulator subroutines access it for data read and write purposes. The CDD-SAT system according to our UML designs, heavily relies on the DB component shown in Fig. 7, and its contents are the actual resultant trials made from Fig. 5 and the CDD-SAT program indeed.

```
1/PhaseContainers:A1,B1,C1,D1,E1,F1,G1,H1,A2,B
2,C2,D2,E2,F2,G2,H2,A3,B3,C3,D3,E3,F3,G3,H3,A4,
B4,C4,D4,E4,F4,G4,H4,A5,B5,C5,D5,E5,F5,G5,H5,A
6,B6,C6,D6,E6,F6,G6,H6,A7,B7,C7,D7,E7,F7,G7,H7;
Alfa:B7;Beta:H4;Gamma:A2;Red:A1,B1,C1,D1,E1,F
1,G1,H1,A2,B2,C2,D2,E2,F2,G2,H2,A3,B3,C3,D3,E3,
F3,G3,H3,A4,B4,E4,F4,G4,H4,F5,G5,H5,F6,G6,H6,G
7,H7;Orange:C4,D4,A5,B5,C5,D5,E5,A6,B6,C6,D6,E
6,A7,B7,C7,D7,E7,F7;Green:;TotalSortingTime:0.6
8;TotalDetectionTime:193.69;
2/PhaseContainers:B1,C1,D1,E1,F1,G1,B2,C2,D2,E
2,F2,G2,B3,C3,D3,E3,F3,G3,B4,C4,D4,E4,F4,G4,B5,C
5,D5,E5,F5,G5,B6,C6,D6,E6,F6,G6;Alfa:B6;Beta:G3;
Gamma:B3;
END
```

**Fig. 7** The raw data output in form of a database file.

Fig. 7, shows that the DB file is saturated with phases' outputs "1/, 2/" headers showing that two simulation phases have been succeeded. The file ends with an "END" line as the final line written by the program. The file is stored as an accessible DB component to supply any information necessary in aim of displaying the outcome of the hybrid simulator. The protocol defining the format of the DB file was customized by the group (programmers) as agreed to communicate under shared points of information, running one CDD-SAT simulation phase from another. Specific *string values* like "1/", "2/",… "Alfa" "Beta", "Gamma", …, "END", are picked up from the file by the program as coded in VB, C++ and Java during system communications. The current component was quite suitable to fulfill the needs of the current release of the program. Thus partitioning this component in the future is essential to manage a huge population of cargo stacks at ports efficiently. More programming is then required to manage such DB files to maintain intelligence when concurrent triangulations for different stacks at different port locations are initiated within the cargo environment.

## 5.3 MODEL

The *domain knowledge* used in the CDD-SAT system is gathered in the simulation-based model. The system makes use of this knowledge when translating raw data gained from the stack labels and SDD tests during the phases of triangulation for the whole cargo population set. As shown in Fig. 5, and as specified in the last row of Table 3, all the maximums and minimum limits of the operations conducted between SAT and SDD systems are defined within the attributes and operations of the model. As all classes get instantiated in VB, C++ or Java within this hybrid simulator, creating timeline charts like Fig. 9, and profiling estimates in form of a DB table like relational-DB Tables 4 and 5, out of the raw data (like the raw DB file in Fig. 7), makes this model robust enough in any hybrid simulation, representing a uniform set of results without confusions to its user.

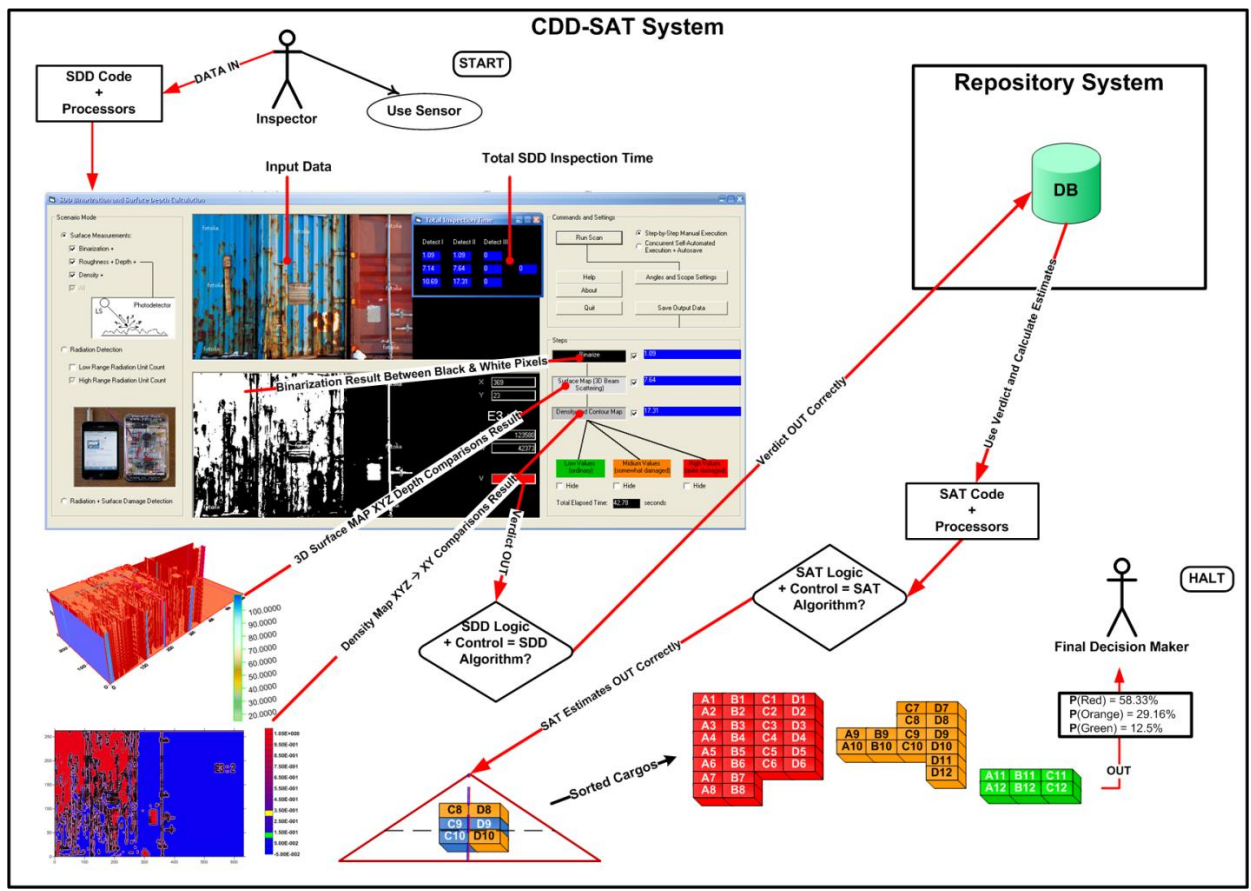

**Fig. 8:** An overview of CDD-SAT results diagram, with key actors, engaging SDD and SAT algorithms with their processors active in the CDD-SAT system.

## 5.4 SIMULATION RESULTS

The main simulation results are recognizable by the DB file component and tables as we have discussed and listed in the previous sections. The final results are at the end of the simulation, profiled to present the cargo status to its manager. Furthermore, the calculated time results are fully explained later in § 6. However, in this section, we show that the collaboration of SDD and SAT is significant when it comes to CDD-SAT system's operations, as well as performance, contributing to an IDSS solution in result.

The IDSS solution is shown in Fig. 8. In this figure, the DB component plays the major role in providing all temporary dataset to the SDD and SAT algorithms, once detection is made and reported to the system by the inspector using other agents (sensors) scanning cargos at the point of START, down to the point of HALT. This is when the manager receives information on cargo status to pass a decision. The diagram also signifies the importance of logic in processing data to supply the manager(s) meaningful data i.e. "information" which further demonstrates the system to be closely tied to a knowledge-based system in generating reliable estimates. The CDD-SAT system in this diagram reiterates the very fundamental design illustrated in Fig. 2, highlighting that SDD with SAT must always be in collaboration with each other between phases to generate accurate estimates to their phases' predecessors.

## 5.5 INTELLIGENCE

The intelligent aspects of our prototype are the functionality of providing suggestions based on the available input. This is an activity usually performed by a skilled human and a set of sensory devices (detectors) identified as agents in our design. However, the system takes into account valid inputs by using our prototype with a certain knowledge level built upon agents' input as well as computations on the same input. The system in turn (which inspection phase?) delivers an output as a suggestion to inspect specific cargo labels in the cargo environment. The inspector no longer has to keep track of all inspection results and does not have to conduct inspection on all cargos. The system handles the relay points between detectors, and _optimally_ runs 3 detections _concurrently_ by the α, β and γ detectors. The inspector and the CDD-SAT system core where a decision is suggested are in position during these steps (recall Use-Case 4.3.3).

An intelligent decision is suggested to inspect a particular container during triangulations to human agents: firstly, the inspectors use portable sensors (detectors), and for quicker results use automated screens moving between containers; secondly, submit profiles to the system manager(s) as a resultant report on cargo inspections; finally, a higher managerial decision is made i.e. *what to do with the damaged version of the cargo*? We have not covered the last one since it involves decision making solutions on a longer timescale during transportation i.e. a tactical model which is out of the current prototype's scope and suitable as an updated version producing sophisticated list of options relative to company protocols. These protocols, depending on what type of e.g. shipping company, are assigned to its employees playing the role of cargo managers, inspectors, etc.

As mentioned before, (§ 5.3), the CDD-SAT Simulation model-base is more of a low-level or a combination of an Operational Model and Real-time Analytical Models which entail mathematical calculations on the written data to the DB file(s). These files are stored for future use on each progressive phase of the SAT algorithm according to scenarios.

Associated requirements for the inspector to scan and suggest which cargo to inspect, is displayed by the program after stack population is inputted. Then, with a simple "Run Scan" button, the user gets a list of 3 containers per phase operation, detecting damages from the undamaged.

From an AI viewpoint, The information gathering part by DB's plays an important role, say, when the same cargo list is suggested with similar damaged vs. undamaged results on the profile, the *program skips* the same configuration of detections (leaving out unnecessary program routines) and gives a high probability to the inspector that what is going to be the outcome before running a formal inspection! For instance, if A2, E2, B2 cargo stacks, turn out to be Red, Orange and Orange in a population of 200 cargo stacks based on previous damage estimates on nearby stacks A1, D1 and B1, we could estimate the same outcome for a *new population input* having the same configuration of damage points. Now, assume that all data points are averaged in terms of their surface damages close to newly-suggested cargos in a larger population $n > 200$. The algorithm, then knows that the outcome of the same stacks in the new population is the same as the one recorded in the past on A2, E2 and B2 for a particular scenario e.g., radiation. The level of exposure and its progress is similar, if the record on these labels were previously stored for the same scenario in the database. So, *the scenario type*, *local configuration of stacks* (its topology) *similar to the previous profiles, most likely gives the similar outcome in SAT operations*.

# 6. Project Optimization

Along the dimensions of *concern* and *uncertainty* of the simulation project, we have studied resource allocation, analysis and time dimensions on cargos using *sensitivity analysis*, with a focus on *optimization*. Based on our *sensitivity quantitative analysis* [6], the model precludes the negative quality of processing information (qualitative) in favor of positive results between detectors installed on the system *excluding human agents* (only considering the software, and hardware detectors). On the other hand, the system is sensitive to the qualitative performance when it comes to servicing the managers with fast data-read relative to delivering reliable estimates on time. Therefore, the *sensitive data* that infer to the actual input data from sensory units (detectors) are tested on a trial basis that could reveal any uncertainties in the inspection process. This exactly complies with the notion stating that "If results are consistent it provides stronger evidence of an effect and of generalizability" (Leamer, 1978) [6]. To this account, the "consistency" is anchored in the time parameter representing a full detection quantified by *time of effort* and processed in a qualitative manner to human agents. The latter must convey to the delivery of information and the estimates made on damaged vs. undamaged cargos, questioning the fact that whether this profile displays reliable estimates (information) to the manager or inspector in a real-time situation. We have computed and compared this aspect with other algorithms along the dimension of time relative to resource allocation for reliable estimates by maintaining performance under different circumstances of cargo inspection:

## 6.1 Saved Time by SAT vs. Time Loss

*Optimization* is gained based on the on-time performance relative to DB information **updates** and **detection** parameters. Here is an example from the main plan on a set of cargo stacks, with a population of $n = 48$:

The equation on the Saved Time Effort from our defined parameters in §3.4 is given by

$$T_{saved} = T_{other} - T_D . \qquad (5)$$

According to the depicted graph, the system operating with SAT for $n = 48$ stacks via 9 inspections takes approximately 113 seconds while for the same population via 48 inspections without SAT (other algorithms) takes approximately $602.6\bar{6}$ seconds, thus, the saved time effort by SAT is $T_{saved} = 486.6\bar{6}$ seconds.

## 6.2 Learnability of the SAT system

The system learns during trials and cargo damage detections. The data is updated regularly: 3-time inspections made over a population of cargos. Of course, the inspections are $3/n$ for each triangulation relative to a red-orange-green estimate. The repeated record by chance during trials of running the SAT application, fairly contributes to AI features explicated earlier in §5.5 of this report.

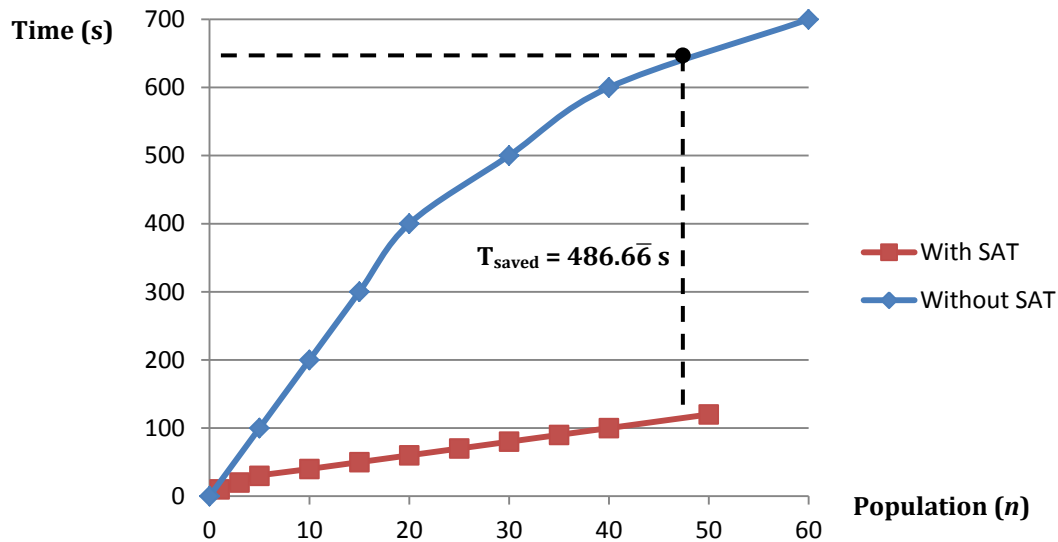

**Fig. 9:** Time performance comparison to other algorithms made by calculating the total saved time deduced from total time calculated on CDD-SAT.

## 7. EVALUATION

### 7.1 VERIFICATION AND VALIDATION

Further analysis to our conducted SWOT analysis, like *sensitivity analysis* on the time and spatial factors (performance) has already been covered. Of course, this resulted in assessing and tackling **Uncertainties** in the project. It further resulted in **Verification and Validation** of our incorporated technique. The most influencing parameter was time raised in the denominator part of our equations i.e. the cargo stack population being processed by the system is measured with respect to time during code executions.

For precise measurements, the length of code execution paths, program volume, difficulty and length, we could use *Halstead complexity measurement* indeed [8]. However, in our hybrid prototype, typical *time functions* were used and summed from one execution step to another. For instance, in VB, we code

```
1.  'The following creates a timer variable
    to compute the elapsed time interval per
    an invoked operation
2.  StartTime = Timer()
3.  ... 'this is where all the code subsists
    and thus timed
4.  EndTime = Timer()    'this is where the
    timer finishes timing
5.  Me.Text1.text = FormatNumber(EndTime -
    StartTime, 2) 'A text box displays the
    elapsed time result down to 2 decimal
    points on the same form object
6.  Debug.Print 'print elapsed time output
```

The last line is then saved or displayed as part of the total elapsed time of the whole SAT operation executed by the other compiled codes in C++ and Java. Then, at the end of all operations of the simulation program, the "total time" T, is measured and compared to the contributed $T_D$ of the program. If T is lost compared to $T_{other}$, we say that SAT acted inefficiently; otherwise, it is always $T_{saved}$, according to our simulation results. Meaning that, the detections and processing lengths are all done and measured in terms of time.

The following graph shows how much time we have totally saved on a random scenario for each inputted cargo population to the system during optimization, without including learnability features as explained in § 6.2. It shows in all scenarios we have saved at least 80 seconds depending on the captured image from the cargo scene, as well as 3D scattering, surface and contour mapping technique with relative activities performed by agents on the containers. The radiation scenario of the project was not covered but considered throughout the design, since it requires a reliable dataset as our sample to be loaded, used, processed and tested by our system, averaging radiation units indicating exposure, thereby generating color-coded red, orange and green results as specified in our design.

We have tested and retested the simulation to conclude our results in form of DB tables and time response measurement. We ought to prove that the SAT algorithm is quite useful in terms of its usability, maintainability, functionality, efficiency, portability and information integrity, based on I/O database and management. For instance, the system user is not overwhelmed with inputting a variety of data to the system (e.g. complicated forms to fill-in) and mainly deals with a simple interface to either relay actual data via sensors on cargo relative to giving an input on just population (passing an argument *n* only) to the system. The rest is the system's duty to process, analyze and update the given information between relay points. The results of the test are quite lucid in ISO-25010 software quality characteristics [7], prior to quantitative analysis, which has been conducted in the first place during the programming phases of the application.

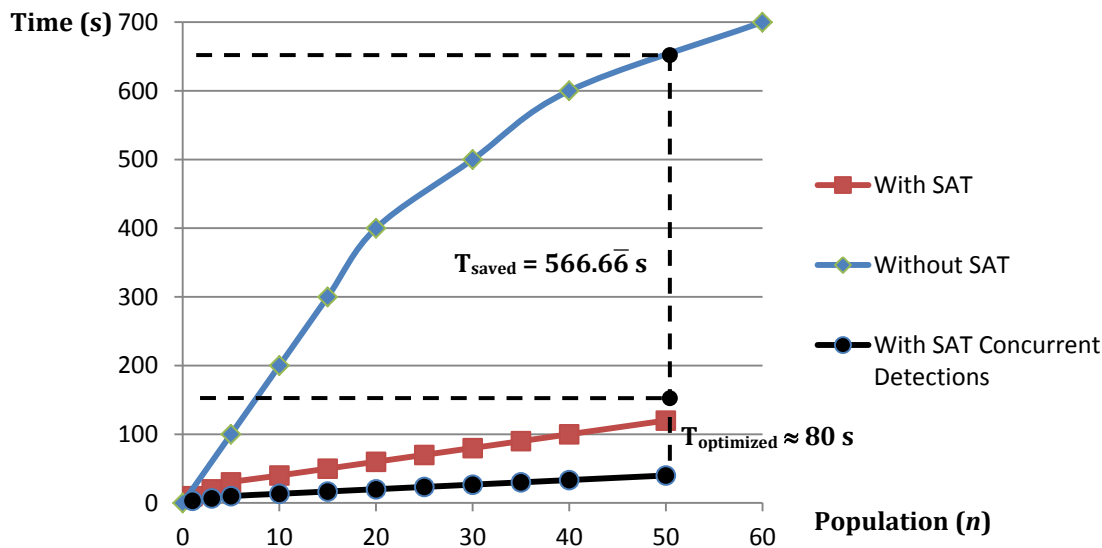

**Fig. 10:** Possible time optimization by running concurrent SDD detections, rather than sequential SDD's.

The 'concurrent scans' curve in Fig. 10, denotes "3 detections by the α, β, γ detectors" which generated approximately $T_D/3$ when the "concurrent scan" option was chosen on the prototype. This shows that compared to a "sequential scanning" by default (running one detector at a time) in the program, a significant time reduction is achieved for better performance. Therefore, Fig. 9 is transformed into Fig. 10, showing this optimization by result.

This is achieved by parallel programming with manageably-running multiple threads on the processor (coded in the detections' simulation part in VB.NET), to manage time sessions or application slices of tasks allocated to process incoming imagery data from the detectors. As a minimum requirement, having a dual-core CPU satisfies this spawning of threads during program executions. Of course, according to ISO 25010, nowadays computers all are *at least* equipped with dual-core CPUs satisfying the usability aspect relative to affordability and cost effective issues within the software engineering quality characteristics [7].

## 8. FUTURE WORK

Pertinent to our analysis, the future work on the CDD-SAT system and its algorithms depend on how we treat our dataset in real-world situations. The current simulation covered certain plausible scenarios portraying a real condition of a cargo relative to environmental changes, impacts, and other constraints which could involve humans in the cargo transportation process.  To this account, we have specified the real scenarios where we can further mimic them and analyze their future perceptions of how they manifest themselves to humans in our real-world relationships, given in our conclusions section of this report.  Of course, the ultimate aim would be to run concurrent processes relative to multiple scenarios of the CDD-SAT application mapping the world to its system topological matrix  (its *dynamic behavior*). That is, a combination of "what-if" analysis for a specific scenario coinciding with another which constitutes a bigger uncertainty predicted by the system. This makes, not only the *reliability* aspect of the system sound, also elevates the *confidence* level of the software to 1.

## 9. CONCLUSIONS

In this project, we have managed to run the simulation on different scenarios. The scenarios were simulated in terms of *xy* local damage estimates relative to *xyz* cargo surface detection scenarios, initially gathered from the cargo environment, once a set of containers are suggested by the CDD-SAT system to the inspector. In particular: probable rusty and oxidized surfaces of containers against undamaged ones, whereas their outcomes are stored in {Red OR Orange OR Green}. This constituted the verdict coming from the system to profile the status of the triangulated cargos. Similarly, probable punctured walls or walls with holes against undamaged ones were also simulated. Concurrent detections processed by the program satisfying 3 incoming detections from the scene for surface damage analysis, proved how fast the system works.

However*, probable radiation exposure was considered but not covered due to in need of trying an actual dataset (to be supplied by our supervisor), which varies in short time intervals in any situation. For instance, based on Roentgen radiation unit counts per minute (cpm), the rate of exposure compared to the rate in other scenarios like rusty container surfaces stationed in some hot-humid environment, is high. The scale of radiation spread to elsewhere is quite evident, whereas the other scenario is deemed to be a long term issue co-occurring between the two scenarios.*

We have animated our Use-Case Diagrams entailing surface detection plus *xy* triangulations on local stack population *n*, where the Outcomes were sorted in {Red OR Orange OR Green} accordingly. This constituted our I/O database results, thus our profile results on the cargo for the whole stack population. Time Length Calculations + Optimization Issues were discussed and show a significant upgrade in making reliable decisions on time compared to other algorithms. To this account, the discussion of the most dominant factors in this IDSS project becomes relevant: stack population, its data acquisition, detection process, and local damage detection sorting with respect to time. In our conclusions, we claim that the CDD-SAT's current usage as an IDSS solution is relevant to "prediction scenarios" based on reliable estimates on damaged vs. undamaged cargos. This makes inspections quite fast at the port and elsewhere. Nevertheless, uncertainties and certain influential factors e.g., cargo environment, human interactions, transportation status, atmospheric impact, etc. always remain visible in our real-world relationships.